\documentclass{article}

\usepackage{PRIMEarxiv}
\usepackage[utf8]{inputenc} 
\usepackage[T1]{fontenc}    
\usepackage{url}            
\usepackage{booktabs}       
\usepackage{amsfonts}       
\usepackage{nicefrac}       
\usepackage{microtype}      
\usepackage{lipsum}
\usepackage{fancyhdr}       
\usepackage{graphicx}       
\usepackage{listings}
\usepackage{xcolor}
\usepackage{float} 
\usepackage{subcaption}
\usepackage{url}
\usepackage{hyperref}       
\usepackage{listings}
\usepackage{xcolor}

\definecolor{myblue}{rgb}{0.13, 0.13, 1.0}
\definecolor{mygreen}{rgb}{0.0, 0.5, 0.0}
\definecolor{myred}{rgb}{0.7, 0.0, 0.0}

\lstdefinestyle{Rstyle}{
  language=R,
  basicstyle=\ttfamily\fontsize{12}{14},
  frame=single,
  keywordstyle=\color{black},
  commentstyle=\color{mygreen},
  stringstyle=\color{myred},
  showstringspaces=false,
  columns=fullflexible,
  xleftmargin=0.4em,
  xrightmargin=0.4em,
}

\graphicspath{{media/}}     

\definecolor{linkcolor}{HTML}{FF0000}    
\definecolor{citecolor}{HTML}{008000}    
\definecolor{urlcolor}{HTML}{0000FF}     

\hypersetup{
    colorlinks=true,
    linkcolor=linkcolor,
    citecolor=citecolor,
    urlcolor=urlcolor,
}

\title{A Comprehensive Guide to Combining R and Python code for Data Science, Machine Learning and Reinforcement Learning
}

\author{
  Alejandro L. García Navarro, Nataliia Koneva, Alfonso S\'{a}nchez-Maci\'{a}n, Jos\'{e} Alberto Hern\'{a}ndez\\
  Departamento de Ingenier\'{i}a Telem\'{a}tica \\
  Universidad Carlos III de Madrid, Spain \\  
  \texttt{\{agnavarr,nkoneva\}@pa.uc3m.es, \{alfonsan,jahgutie\}@it.uc3m.es} \\
}

\begin{document}
\maketitle

\begin{abstract}
Python has gained widespread popularity in the fields of machine learning, artificial intelligence, and data engineering due to its effectiveness and extensive libraries. R, on its side, remains a dominant language for statistical analysis and visualization. However, certain libraries have become outdated, limiting their functionality and performance. Users can use Python's advanced machine learning and AI capabilities alongside R's robust statistical packages by combining these two programming languages. This paper explores using R's \texttt{reticulate} package to call Python from R, providing practical examples and highlighting scenarios where this integration enhances productivity and analytical capabilities. With a few hello-world code snippets, we demonstrate how to run Python's scikit-learn, pytorch and OpenAI gym libraries for building Machine Learning, Deep Learning, and Reinforcement Learning projects easily. 
\end{abstract}


\section{Introduction}

In recent years, data science and machine learning fields have experienced a rise in the use of Python and R \cite{raschka2020machinelearningpythonmain, castro2023landscapehighperformancepythondevelop}. Python is often regarded as a tool with the greatest amount of libraries and tools designed for machine learning, artificial intelligence, and data engineering. Conversely, R remains a go-to language for statistical analysis and advanced visualization, thanks to packages along the lines of \textit{stats} \cite{stats}, \textit{caret} \cite{caretpackage}, \textit{ggplot2} \cite{ggplot2} or \textit{shiny} \cite{shiny}. 

In the evolving landscape of data science, combining multiple programming languages has become a popular strategy to take advantage of the strengths of each. For example, research has explored integrating Julia and Python for scientific computing to use Julia's computational efficiency alongside Python~\cite{osborne2024bridgingworldsachievinglanguage}. Similarly, the integration of Stata and Python has been examined to enhance machine learning applications, as shown in~\cite{cerulli2021machinelearningusingstatapython}, which details how Stata's recent integration with Python allows for optimal tuning of machine learning models using Python's scikit-learn library.

In the AI/ML field, it often happens that many libraries and open-source code examples appear in Python, and it takes some time until libraries are ported to R. Therefore, R programmers have it difficult to start using new AI/ML libraries developed in Python. On the other hand, R is well-known to provide thousands of libraries dedicated to statistical analysis and tests, many of them not yet written in Python. 

This paper explores the \texttt{reticulate} package \cite{reticulate}, which acts as a bridge between R and Python, allowing programmers to use both languages within a single workflow. This makes it easier for programmers to combine Python's cutting-edge machine-learning capabilities with R's statistical tools, hence creating a more versatile environment. By showing practical examples with code snippets, we aim to demonstrate how to combine the best of both worlds. Such hello-world code snippets show clear examples for using classical Machine Learning, Deep Learning and Reinforcement Learning libraries like \texttt{scikit-learn}, \texttt{pytorch} and \texttt{OpenAI Gym}. 

The structure of the paper is as follows: Section \ref{sec:package} provides an introduction to the package, highlighting its key features and walking through the installation process. Section \ref{sec:cases} presents some examples, including code snippets; and finally, Section \ref{sec:conclusions} concludes the paper.

\section{The reticulate Package}
\label{sec:package}
\subsection{Introduction to reticulate}
The \texttt{reticulate} package is a tool that allows users to call Python from R and R from Python, combining the strengths of both programming languages in a single workflow. In this paper, we will focus solely on calling Python code from an R environment.

Reticulate allows to import any Python module and access its functions, classes, and objects from R, enabling a more versatile and flexible approach to data analysis, machine learning, and statistical computing.

\subsection{Key features and functionalities}
Combining these two languages brings together the best of both worlds:
\begin{enumerate}
    \item Choose the best tool for each task by using R’s statistical analysis and Python’s programming and machine learning strengths. 
    \item Access more libraries and packages from both ecosystems. 
    \item Easy transfer of data between R and Python for flexible data handling in complex analysis pipelines.
\end{enumerate}

\subsection{Installation and setup}
Before using the library, we have to make sure that we have the following prerequisites:
\begin{enumerate}
    \item R Installation: \href{https://cran.r-project.org/}{Install R} on your system.
    \item Python Installation: \href{https://www.python.org/downloads/}{Install Python} on your system.
    \item RStudio (Optional but recommended): Using RStudio as your IDE can simplify the process of using \texttt{reticulate}. Download RStudio from \href{https://posit.co/download/rstudio-desktop/}{here}.
\end{enumerate}

Once you have completed all the prerequisites, it is time to install the package. Use the following command in your R console:
\begin{lstlisting}[style=Rstyle]
if (!require("reticulate")) install.packages("reticulate")
\end{lstlisting}

After installation, you need to load the package using:
\begin{lstlisting}[style=Rstyle]
library(reticulate)
\end{lstlisting}

\section{Practical Implementation}
\label{sec:cases}

\subsection{Basic Usage}

\subsubsection{Importing Python Modules}
To import a Python module in R using the \texttt{reticulate} package, you use the \textit{import} function. For example, to import the NumPy library \cite{harris2020array}, you can use:
\begin{figure}[H]
\centering
\begin{lstlisting}[style=Rstyle]
np <- import("numpy")
\end{lstlisting}
\caption{Importing the NumPy library in R using reticulate.}
\end{figure}

With this, you can use the \textit{np} object to access NumPy functions and methods just as you would in Python:
\begin{figure}[H]
\centering
\begin{lstlisting}[style=Rstyle]
array <- np$array(c(1, 2, 3, 4, 5))
\end{lstlisting}
\caption{Creating a NumPy array in R using reticulate.}
\end{figure}

\subsubsection{Running Python Code in R}
Sometimes, it might be useful to execute Python code directly within an R script, and this can be easily done using the \textit{py\_run\_string} function. This function allows you to run Python code as a string:
\begin{figure}[H]
\centering
\begin{lstlisting}[style=Rstyle]
py_run_string("print('Hello from Python')")
\end{lstlisting}
\caption{Executing Python code directly within an R script using \textit{py\_run\_string}.}
\end{figure}

Alternatively, it may be more convenient to directly execute a Python script file. For this, you can use the \textit{py\_run\_file} function:
\begin{figure}[H]
\centering
\begin{lstlisting}[style=Rstyle]
# py_run_file("path/to/your_script.py")
py_run_file("test.py")
\end{lstlisting}
\caption{Executing a Python script file from R using \textit{py\_run\_file}.}
\end{figure}

\subsubsection{Accessing Python Objects in R}
In the same way, you can access and manipulate Python objects in R. For example, if you create a Python list, you can access it in R:
\begin{figure}[H]
\centering
\begin{lstlisting}[style=Rstyle]
# You can access a Python list
py_run_string("my_list = [1, 2, 3, 4, 5]")
my_list <- py$my_list
print(my_list)

# You can manipulate the list
my_list[1] <- 4
print(my_list)
\end{lstlisting}
\caption{Accessing and manipulating a Python list in R.}
\end{figure}

It is also possible to define functions and call them from R:
\begin{figure}[H]
\centering
\begin{lstlisting}[style=Rstyle]
py_run_string("
def greet(name):
  return 'Hello, ' + name + '!'
")

greet <- py$greet
print(greet("World"))
print(greet("James"))
\end{lstlisting}
\caption{Defining and calling a Python function from R.}
\end{figure}

\subsection{Data Manipulation}

\subsubsection{Using Python Libraries Like NumPy and pandas}
You can use Python libraries like NumPy and pandas \cite{mckinney2010data} for data manipulation in R:
\begin{figure}[H]
\centering
\begin{lstlisting}[style=Rstyle]
# Import NumPy and pandas
np <- import("numpy")
pd <- import("pandas")

# Create a numpy array
array <- np$array(c(1, 2, 3, 4, 5))
print(array)

# Create a pandas data frame
py_df <- pd$DataFrame(dict(a=np$array(c(1, 2, 3)), b=np$array(c('x', 'y', 'z'))))
print(py_df)
\end{lstlisting}
\caption{Using Python libraries like NumPy and pandas for data manipulation in R.}
\end{figure}

\subsubsection{Converting Data Types Between R and Python}
It is important to know that the \texttt{reticulate} package automatically converts many data types between R and Python. For example:
\begin{enumerate}
    \item R vectors become Python lists.
    \item R data frames become pandas data frames.
\end{enumerate}

You can manually convert data types using specific functions if needed:
\begin{enumerate}
    \item To convert an R data frame to a pandas data frame:
    \begin{lstlisting}[style=Rstyle]
    # Define data frame
    df <- data.frame(a = 1:3, b = c('x', 'y', 'z')) 
    
    # Convert R data frame to pandas data frame
    py_df <- r_to_py(df)
    print(py_df)
    \end{lstlisting}
    \begin{figure}[H]
    \centering
    \caption{Converting an R data frame to a pandas data frame.}
    \end{figure}

    \item To convert a pandas data frame back to an R data frame:
    \begin{lstlisting}[style=Rstyle]
    # Convert pandas data frame to R data frame
    r_df <- py_to_r(py_df)
    print(r_df)
    \end{lstlisting}
    \begin{figure}[H]
    \centering
    \caption{Converting a pandas data frame back to an R data frame.}
    \end{figure}
\end{enumerate}

\subsection{Visualization}

\subsubsection{Using Python Visualization Libraries}
Sometimes, you might have more experience plotting in Python than in R. Thanks to this package, Python libraries like Matplotlib \cite{Hunter:2007} and Seaborn \cite{Waskom2021} can be used:
\begin{figure}[H]
\centering
\begin{lstlisting}[style=Rstyle]
# Import libraries
plt <- import("matplotlib.pyplot")
sns <- import("seaborn")

# Create a plot using Matplotlib
plt$figure()
plt$plot(c(1, 2, 3), c(4, 5, 6))
plt$show()

# Create a plot using Seaborn
plt$figure()
sns$set_theme()
df <- sns$load_dataset("iris")
sns$scatterplot(data=df, x="sepal_length", y="sepal_width", hue="species")
plt$show()
\end{lstlisting}
\caption{Creating plots using Matplotlib and Seaborn in R with reticulate.}
\end{figure}

In the images below, you can see the two different plot types created using Matplotlib and Seaborn. The line plot on the left demonstrates a simple linear relationship, while the scatter plot on the right provides a detailed view of the Iris dataset, highlighting the differences in sepal dimensions across species:
\begin{figure}[h]
    \centering
    \begin{subfigure}[b]{0.45\textwidth}
        \centering
        \includegraphics[width=\textwidth]{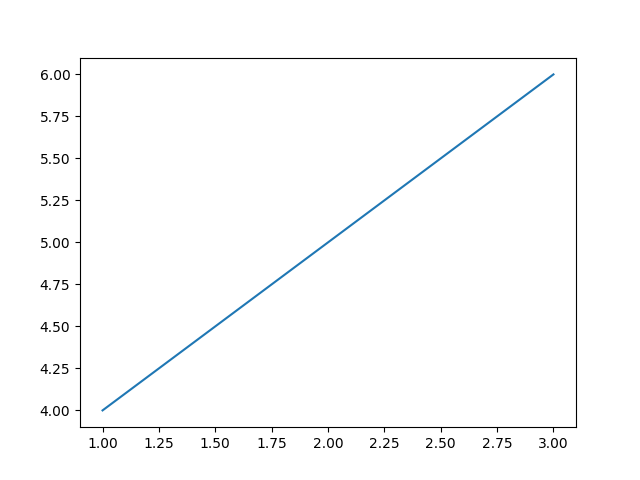}
        \caption{Line Plot}
    \end{subfigure}
    \hfill
    \begin{subfigure}[b]{0.45\textwidth}
        \centering
        \includegraphics[width=\textwidth]{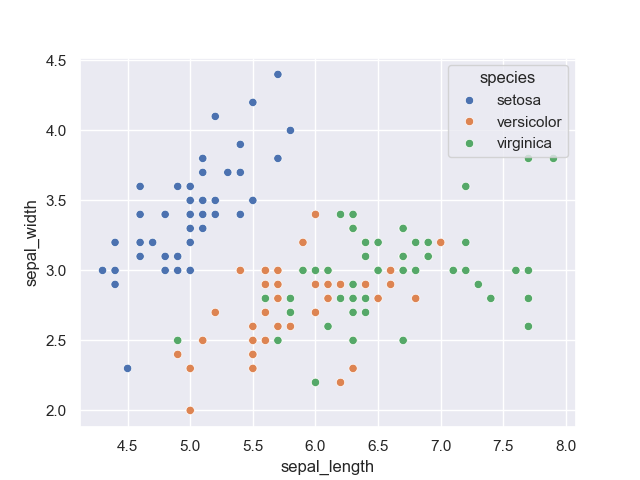}
        \caption{Scatter Plot}
    \end{subfigure}
    \caption{Output of Line and Scatter Plots.}
    \label{fig:plots}
\end{figure}

\subsection{Machine Learning}

\subsubsection{Using Scikit-Learn for Classification}
In some cases, you might want to perform classification tasks using Scikit-Learn \cite{scikit-learn} in R. This example shows how to import the necessary libraries, load the Iris dataset, and train a Support Vector Classifier (SVC) model:
\begin{figure}[H]
\centering
\begin{lstlisting}[style=Rstyle]
# Import scikit-learn
sklearn <- import("sklearn")
datasets <- sklearn$datasets
svm <- sklearn$svm
metrics <- sklearn$metrics

# Load dataset and train a model
iris <- datasets$load_iris()
X <- iris$data
y <- iris$target
model <- svm$SVC()
model$fit(X, y)

# Make predictions
predictions <- model$predict(X)

# Evaluate the model
accuracy <- metrics$accuracy_score(y, predictions)
print(paste("Accuracy:", accuracy))
\end{lstlisting}
\caption{Using Scikit-Learn for classification in R.}
\end{figure}

\subsubsection{Building and Evaluating a Regression Model}
You can also build and evaluate a regression model using Scikit-Learn. This example demonstrates how to import the necessary libraries, load the diabetes dataset, split it into training and testing sets, and train a linear regression model:

\begin{figure}[H]
\centering
\begin{lstlisting}[style=Rstyle]
# Import necessary libraries
sklearn <- import("sklearn")
datasets <- sklearn$datasets
linear_model <- sklearn$linear_model
metrics <- sklearn$metrics

# Load the diabetes dataset
diabetes <- datasets$load_diabetes()
X <- diabetes$data
y <- diabetes$target

# Split the data into training and testing sections
library(zeallot)
train_test_split <- sklearn$model_selection$train_test_split
c(X_train, X_test, y_train, y_test) %<-% train_test_split(X, y, test_size = 0.2)

# Train a linear regression model
model <- linear_model$LinearRegression()
model$fit(X_train, y_train)

# Make predictions
predictions <- model$predict(X_test)

# Evaluate the model
mse <- metrics$mean_squared_error(y_test, predictions)
print(paste("Mean Squared Error:", mse))
\end{lstlisting}
\caption{Building and evaluating a regression model using Scikit-Learn in R.}
\end{figure}

\subsubsection{Using PyTorch for Deep Learning}
PyTorch \cite{paszke2019pytorchimperativestylehighperformance} is another popular deep learning framework that can be used within R via the \texttt{reticulate} package. This example shows how to import PyTorch, define a neural network, and train it on the MNIST dataset.

Due to the extensiveness of the code, this section is explained in steps:
\begin{enumerate}
    \item Import necessary modules
    \begin{lstlisting}[style=Rstyle]
    torch <- import("torch")  
    torchvision <- import("torchvision")  
    nn <- torch$nn  
    optim <- torch$optim  
    transforms <- torchvision$transforms    
    \end{lstlisting}
    \begin{figure}[H]
    \centering
    \end{figure}

    \item Define hyperparameters
    \begin{lstlisting}[style=Rstyle]
    batch_size <- as.integer(64)  # Number of samples per batch
    learning_rate <- 0.001  # Learning rate for the optimizer
    num_epochs <- 5  # Number of times to iterate over the entire training dataset
    \end{lstlisting}
    \begin{figure}[H]
    \centering
    \end{figure}
    
    \item Data preparation
    \begin{lstlisting}[style=Rstyle]
    # Define a transformation to convert images to tensors and normalize them
    transform <- transforms$Compose(list(transforms$ToTensor(), 
    transforms$Normalize(c(0.5), c(0.5))))
    
    # Load the training and test datasets with the defined transformations
    train_dataset <- torchvision$datasets$MNIST(root='./data', train=TRUE, 
    transform=transform, download=TRUE)
    train_loader <- torch$utils$data$DataLoader(dataset=train_dataset, 
    batch_size=batch_size, shuffle=TRUE)
    test_dataset <- torchvision$datasets$MNIST(root='./data', train=FALSE,
    transform=transform, download=TRUE)
    test_loader <- torch$utils$data$DataLoader(dataset=test_dataset,
    batch_size=batch_size, shuffle=FALSE)
    \end{lstlisting}
    \begin{figure}[H]
    \centering
    \end{figure}

    \item Define neural network model using Python code
    \begin{lstlisting}[style=Rstyle]
    py_run_string("
    import torch
    import torch.nn as nn
    import torch.nn.functional as F
    
    # Define a simple neural network
    class SimpleNN(nn.Module):
        def __init__(self):
            super(SimpleNN, self).__init__()
            # First fully connected layer (input layer)
            self.fc1 = nn.Linear(28*28, 512)
            # Second fully connected layer (hidden layer)
            self.fc2 = nn.Linear(512, 256)
            # Third fully connected layer (output layer)
            self.fc3 = nn.Linear(256, 10)
        
        # Define the forward pass
        def forward(self, x):
            x = x.view(-1, 28*28)  # Flatten the input image
            x = F.relu(self.fc1(x))  # Apply ReLU activation to the first layer
            x = F.relu(self.fc2(x))  # Apply ReLU activation to the second layer
            x = self.fc3(x)  # Output layer
            return x
            
    model = SimpleNN()
    ")
    
    # Access the model defined in the Python code
    model <- py$model
    
    # Define loss function and optimizer
    criterion <- nn$CrossEntropyLoss()  
    optimizer <- optim$Adam(model$parameters(), lr=learning_rate) 
    \end{lstlisting}
    \begin{figure}[H]
    \centering
    \end{figure}

    \item Train the model
    \begin{lstlisting}[style=Rstyle]
    for (epoch in 1:num_epochs) {
      # Iterate over batches of data
      for (batch in reticulate::iterate(train_loader)) {
        images <- batch[[1]]  # Input images
        labels <- batch[[2]]  # True labels
    
        # Forward pass: compute predicted outputs
        outputs <- model$forward(images)
        # Compute loss
        loss <- criterion(outputs, labels)
    
        # Backward pass: compute gradients
        optimizer$zero_grad()  # Clear existing gradients
        loss$backward()  # Compute gradients
        optimizer$step()  # Update model parameters
    
      }
      cat(sprintf('Epoch [%d/%d], Loss: %.4f\n', epoch, num_epochs, loss$item()))
    }
    \end{lstlisting}
    \begin{figure}[H]
    \centering
    \end{figure}

    \item Test the model 
    \begin{lstlisting}[style=Rstyle]
    model$eval()  
    correct <- 0  
    total <- 0  
    
    # Iterate over batches of test data
    for (batch in reticulate::iterate(test_loader)) {
      images <- batch[[1]]  
      labels <- batch[[2]]  
      outputs <- model$forward(images)  
      predicted <- torch$max(outputs$data, dim=1L)$indices  
      total <- total + as.integer(labels$size(as.integer(0)))  
      correct <- correct + as.integer((predicted == labels)$sum()$item())  
    }
    
    # Compute accuracy
    accuracy <- 100 * correct / total
    cat(sprintf('Accuracy of the model on the 10000 test images: %.2f%%\n', 
    accuracy))
    \end{lstlisting}
    \begin{figure}[H]
    \centering
    \caption{Using PyTorch for deep learning in R.}
    \end{figure}
\end{enumerate}

\subsubsection{Using TensorFlow for Deep Learning}
In case the user lacks experience with PyTorch, we can perform deep learning tasks with TensorFlow \cite{tensorflow2015-whitepaper} and Keras \cite{chollet2015keras}. We will load and preprocess the MNIST dataset, build a neural network model, and train it:\\
\begin{minipage}{\textwidth}
\begin{figure}[H]
\centering
\begin{lstlisting}[style=Rstyle]
# Import TensorFlow and Keras to create and tranii the neural network
tf <- import("tensorflow")
keras <- import("keras")

# Load and preprocess data
mnist_data <- keras$datasets$mnist$load_data()
train_images <- mnist_data[[1]][[1]]
train_labels <- mnist_data[[1]][[2]]
test_images <- mnist_data[[2]][[1]]
test_labels <- mnist_data[[2]][[2]]

# Normalize pixel values
train_images <- train_images/255
test_images <- test_images/255

# Build the model
model <- keras$Sequential() 
model$add(keras$layers$Flatten(input_shape = as.integer(c(28, 28))))
model$add(keras$layers$Dense(units = as.integer(128), activation = "relu"))
model$add(keras$layers$Dense(units = as.integer(10), activation = "softmax")) 

# Compile the model (prepare it for training)
model$compile(optimizer = "adam", loss = "sparse_categorical_crossentropy", 
metrics = list("accuracy"))

# Train the model
model$fit(train_images, train_labels, epochs = as.integer(5), verbose = 0)

# Evaluate the model
metrics <- model$evaluate(test_images, test_labels, verbose = 0)
test_loss <- metrics[[1]]
test_acc <- metrics[[2]]
print(paste("Test accuracy:", test_acc))
\end{lstlisting}
\caption{Using TensorFlow for deep learning in R.}
\end{figure}
\end{minipage}

\newpage
\subsection{Reinforcement Learning}
\subsubsection{Using OpenAI Gym for Reinforcement Learning}
If you are exploring reinforcement learning, OpenAI Gym \cite{brockman2016openaigym} is a great tool that can be used in R. This example shows how to import the necessary libraries, create the CartPole-v1 environment, and run a series of episodes:\\
\begin{minipage}{\textwidth}
\begin{figure}[H]
\centering
\begin{lstlisting}[style=Rstyle]
# Import necessary libraries
gym <- import("gym")
np <- import("numpy")

# Create the environment
env <- gym$make("CartPole-v1")

# Initialize a list to store total rewards per episode
total_rewards <- numeric(50)

# Run episodes
for (episode in 1:50) {
  state <- env$reset()
  total_reward <- 0
  
  for (step in 1:100) {
    # Take a random action
    action <- env$action_space$sample()
    
    # Perform the action in the environment
    result <- env$step(action)
    new_state <- result[[1]]
    reward <- result[[2]]
    done <- result[[3]]
    
    # Accumulate the reward
    total_reward <- total_reward + reward
    
    # Update the state
    state <- new_state
    
    # Break the loop if the episode is finished
    if (done) {
      break
    }
  }
  total_rewards[episode] <- total_reward
  print(paste("Episode:", episode, "Total Reward:", total_reward))
}
env$close()
\end{lstlisting}
\caption{Using OpenAI Gym for reinforcement learning in R.}
\end{figure}
\end{minipage}

\subsection{Web Scraping}
\subsubsection{Using BeautifulSoup for Web Scraping}
This final last section of this paper covers web scraping. It allows us to extract data from websites, and Python's BeautifulSoup library \cite{Richardson2019} is a powerful tool for this purpose. By using it within R, we can easily scrape and analyze web data:
\begin{figure}[H]
\centering
\begin{lstlisting}[style=Rstyle]
# Import BeautifulSoup and requests
bs4 <- import("bs4")
requests <- import("requests")

# Fetch webpage content
url <- "https://www.uc3m.es/Home"
response <- requests$get(url)

# Parse HTML content
soup <- bs4$BeautifulSoup(response$content, "html.parser")

# Extract all div elements
divs <- soup$find_all("div")

# Print the content of the first 10 divs
for (i in 1:10) {
  div <- divs[[i]]
  print(div$get_text())
}
\end{lstlisting}
\caption{Using BeautifulSoup for web scraping to find all div elements.}
\end{figure}

\section{Summary and Discussions}
\label{sec:conclusions}
In this article, we have explored the integration of two programming languages using R's \texttt{reticulate} package, demonstrating its utility across different areas such as data scrapping, manipulation, visualization, and machine learning. From this guide, we can take an important insight: \texttt{reticulate} offers data scientists and researchers a powerful tool to strengthen their analytical workflows by combining the benefits and libraries of R and Python. The examples provided showed how users can import Python modules, execute Python code within R, and manipulate data structures.

We highlighted the implementation of popular Python libraries like NumPy, Pandas, Matplotlib, Seaborn, Scikit-Learn, Pytorch and TensorFlow within R, illustrating how \texttt{reticulate} can be used to develop and evaluate complex machine learning models and visualize data effectively. Additionally, the application of reinforcement learning using OpenAI Gym within the R environment demonstrated that it is even possible to execute reinforcement learning experiments.

In conclusion, this paper remains a didactic guide that shows the combination of R and Python, enabling users to use the best features of both languages. We encourage data practitioners to incorporate \texttt{reticulate} into their workflows to achieve greater insights, using the combined power of R and Python to tackle complex data challenges.

\section*{Acknowledgments}
The authors would like to acknowledge the support of Spanish projects ITACA (PDC2022-133888-I00) and 6G-INTEGRATION-3 (TSI-063000-2021-127) and EU project SEASON (grant no. 101096120).

\bibliographystyle{unsrt}
\bibliography{references}

\end{document}